\documentclass[10pt,journal,compsoc]{IEEEtran}
\usepackage{booktabs}
 \usepackage{graphicx}
 \usepackage[justification=justified]{caption}
 \usepackage{cite}
\usepackage{bm}
\usepackage[T1]{fontenc}
\usepackage{algpseudocode}
\usepackage{algorithm}
\usepackage{amsmath,amsthm,amssymb,amsfonts}
\usepackage{bigstrut}
\usepackage{multirow}
\usepackage{color}
\usepackage{bbding}
\usepackage{pifont}
\usepackage{comment}
\usepackage{wasysym}

\begin{document}
%
\title{Efficient Task Offloading Algorithm for Digital Twin in Edge/Cloud Computing Environment }
%
%
%

\author{Ziru ZHANG,
        Xuling ZHANG,
        Guangzhi ZHU,
        Yuyang WANG,~\IEEEmembership{}
        and Pan HUI,~\IEEEmembership{Fellow,~IEEE}
\IEEEcompsocitemizethanks{\IEEEcompsocthanksitem Ziru Zhang, Xuling Zhang, Guangzhi Zhu, Yuyang Wang, and Pan Hui are with the Information Hub, Hong Kong University of Science and Technology (Guangzhou), China, 511453. \protect\\
E-mail: \{zzhang758, xzhang659, gzhu305\}@connect.hkust-gz.edu.cn, \{yuyangwang, panhui\}@ust.hk
\IEEEcompsocthanksitem This work is supported by Science and Technology Bureau of Nansha District, Grant No.2022ZD012.}}
\IEEEtitleabstractindextext{%
\begin{abstract}
In the era of Internet of Things (IoT), Digital Twin (DT) is envisioned to empower various areas as a bridge between physical objects and the digital world. Through virtualization and simulation techniques, multiple functions can be achieved by leveraging computing resources. In this process, Mobile Cloud Computing (MCC) and Mobile Edge Computing (MEC) have become two of the key factors to achieve real-time feedback. However, current works only considered edge servers or cloud servers in the DT system models. Besides, The models ignore the DT with not only one data resource. In this paper, we propose a new DT system model considering a heterogeneous MEC/MCC environment. Each DT in the model is maintained in one of the servers via multiple data collection devices. The offloading decision-making problem is also considered and a new offloading scheme is proposed based on Distributed Deep Learning (DDL). Simulation results demonstrate that our proposed algorithm can effectively and efficiently decrease the system's average latency and energy consumption. Significant improvement is achieved compared with the baselines under the dynamic environment of DTs.

\end{abstract}

}

\maketitle

\IEEEdisplaynontitleabstractindextext

%
\IEEEpeerreviewmaketitle


\section{\scshape Introduction}

\IEEEPARstart{W}{I}{T}{H} the rapid development of digital technology, DT has triggered significant advancement and interaction in both academia and industry \cite{DT_survey_03,xiao2023vr}. DT, a virtual model designed to accurately represent a physical object, can be regarded as consisting of three main parts: a physical object, its virtual twin, and a mapping between the physical object and its virtual twin \cite{DT_survey_01}. As an emerging and attractive technology, the DT maps the states of physical objects into the virtual world through virtual descriptions or digital representations, significantly reshaping the design and engineering process \cite{DT_survey_02}. In addition, by continuously collecting and analyzing sensor data, DT can model, simulate, monitor, analyze, and optimize the physical world and help people predict the related states of the near future. In recent years, DT technology has been used in various areas, such as real-time remote monitoring, 6G networks, control in industry, MCC, and MEC. From a technical aspect, the DT system comes with a complex architecture to satisfy the functions of virtual-real mapping, real-time synchronization, and prediction \cite{DT_survey_04}. Therefore, the infrastructure of DTs not only needs to provide an arithmetic power guarantee but also needs network transmission equipment that can achieve negligible transmission latency across devices~\cite{wang2022re}.

To address the limitations of the computing power of the DT infrastructure, we expect to take advantage of MCC by offloading complex computing tasks from mobile devices to a central cloud. By leveraging the rich virtual resources and exploiting cloud servers' computational power, we can reduce the stress on mobile devices to handle tasks locally, reducing the task response time. However, when mobile devices communicate with MCC servers over long distances, this approach will be reformulated for many problems, such as high latency, low bandwidth, and network congestion. Hence, offloading tasks and data to MCC servers frequently is infeasible. Researchers have focused their research on MEC to solve the problems encountered by MCC servers. By utilizing the computing resources around the user, MEC provides end-users with powerful arithmetic power, storage capacity, energy efficiency, etc., and enhances the capabilities of MCC servers. Furthermore, the proximity of MEC servers to mobile users considerably shrinks the communication cost of task offloading and significantly reduces network latency. Moreover, MEC is equipped with better offloading technology to meet the increasing demands of ultra-high bandwidth and ultra-low network latency needs \cite{DT_survey_05}. However, when MEC systems deal with complex scenarios, generating offloading decisions using traditional optimization methods or mathematical models consumes much computation time and energy. Therefore, making optimal offloading decisions with less computation time and energy has been a significant challenge.

Artificial Intelligence (AI) algorithms have emerged as a promising approach to the above-mentioned challenge. However, as most machine learning algorithms require large amounts of data to train the network, optimal data for offloading decisions are difficult to obtain in a dynamic and heterogeneous MEC/MCC environment. Combining Reinforcement Learning (RL) with Deep Neural Network (DNN), Deep Reinforcement Learning (DRL) can learn to solve complex problems through trial and error. In addition, DRL can train the network without data. Therefore, researchers have investigated various offloading algorithms for MEC tasks based on DRL. These methods treat the MEC system as a fixed environment and generate optimal offloading decisions without training data. In addition to DRL, DDL has become another promising method to generate near-optimal decisions \cite{Distributed_01}. DDL uses multiple parallel DNNs to generate offloading decisions for each task, which is independent of the training data. Besides, the employment of the DNNs can also enable the model to handle more sophisticated environments and tasks.

It is envisioned that DT, MCC, MEC and AI are critical technologies in the era of IoT. Edge computing and cloud computing technologies are necessary for achieving real-time data synchronization. Furthermore, merging DTs with machine learning algorithms will bring great benefits in improving the utility of the servers. Besides, AI is one of the underlying core technologies of DT. AI algorithms can not only provide data processing and system optimization for DT systems but also shorten the task offloading decision time. However, since DT technologies are still developing, research on the collaboration between DT and AI has yet to mature. Specifically, very little work has explored applying DT and DDL to MEC and MCC.

Based on the existing work, we notice that current DT systems involved either cloud servers or edge servers in the model. But cloud servers and edge servers have different advantages. So, we want to design a new DT system that can integrate DTs with cloud servers and edge servers. With the help of MEC and MCC, DTs can be maintained in the proper server to minimize the average latency of the system. Besides, for the system with multiple DTs and multiple servers, we also want to find an intelligent way to allocate the resources to further utilize the servers in the system. Moreover, most DT systems assume that each DT is linked to one data resource. However, DT with multiple data resources should also be considered, such as Internet of Vehicles (IoV) and Metaverse \cite{6G_Metaverse}.

In this paper, we first propose a new DT system model based on cloud computing and edge computing technologies. To better utilize the resources and decrease the system cost, such as average latency and energy consumption, the resource allocation problem is formalized into an optimization problem. Then we present a new offloading algorithm to the optimization problem based on DDL. Finally, the proposed model and algorithm are evaluated via simulation. The main contributions of this paper can be summarized as follows:

\begin{itemize}
\item We design a new DT system model with one cloud server, multiple edge servers and various DTs. Each DT is maintained in one server using data from multiple resources. The data for the same DT are synchronized in the corresponding server.

\item We formulate the resource allocation problem for placing all the DTs in cloud servers or edge servers according to different scenarios. The problem is formulated as a Mixed-Integer Programming (MIP) problem. By solving the problem, we can lower the system cost and provide a better experience for the users.

\item  Considering the dynamic environment, we design a distributed framework based on DDL to minimize the system cost by allocating server resources intelligently. A fully connected network layer is also added to accelerate the training process and improve portability by further extracting the feature of each group. The DDL model treats features as input rather than the data from all the entities.

\item To evaluate the effectiveness of the proposed algorithm, we conduct several experiments by simulating real MEC/MCC scenarios. A test dataset consisting of 1024 different test data is generated randomly, and the average system costs under several offloading schemes are tested. The numerical results indicate that our method can significantly reduce the system cost.
\end{itemize}

The remainder of this paper is organized as follows. In Section~\ref{related_work}, we summarize related work about DT and offloading algorithm. Next, we evaluate the system model and problem formulation in Section~\ref{system_model}. In section~\ref{algorithm}, we present the DDL driven decision-making algorithm. Then we discuss the simulation process and evaluation result in Section~\ref{evaluation}. In Section~\ref{conclusion}, we conclude this article.

\section{Related Work}
\label{related_work}

The application of DTs has been found in several fields and has achieved significant progress, e.g., IoT, IoV and Unmanned Aerial Vehicle (UAV). To improve the system performance and enhance the stability, researchers also proposed various DT system models according to diverse application scenarios and transmission environments. Besides, cloud computing and edge computing technologies are also widely applied in different models to reduce the latency and energy consumption of the system. Different algorithms are also proposed to find proper offloading decisions. In this section, we will briefly summarize some previous studies about different DT system models and state-of-the-art offloading decision approaches.

\subsection{Digital Twin}

With the big success of AI, it is possible to achieve more sophisticated, intelligent functions on various mobile devices such as drones, self-driving cars, and robots. DT technology, as a bridge between physical entities and virtual systems, has been used in many different areas.

Industrial IoT is an important concept for Industry 4.0. To further improve the performance of the IoT devices in the industry, DT technologies have been adopted in many works. For example, Sun et al.\cite{DT_01} proposed an architecture based on DTs to assist the federated learning process of the Industrial IoT scenario. Song et al. \cite{DT_02} designed a federated learning framework for the DT driven Industrial IoT, where DRL and Lyapunov dynamic deficit queue are used to improve the communication efficiency. Therefore, data security and training accuracy of federal learning are guaranteed with the help of DT. In addition to Industrial IoT, IoV is an important IoT extension. DTs can provide a digital simulation model of the vehicles and optimize the allocation of resources. For instance, Wang et al. \cite{DT_03} and Sun et al. \cite{DT_04} used dynamic DT of aerial-assisted IoV to handle the unified resource scheduling and allocation problem. Alternating Direction Method of Multipliers (ADMMs) and Stackelberg game theory are employed to improve the overall efficiency of the model. Besides, Sun et al. \cite{DT_05} designed a dynamic DT and federated learning driven system for the air-ground networks: different incentive mechanisms are applied to improve the accuracy and energy efficiency of the system. Non-Orthogonal Multiple Access (NOMA) is an important part of next-generation wireless communications. Hence, Wang et al. \cite{DT_06} developed an energy-efficient DT system based on NOMA transmission. All the devices collaborate via federated learning to update a universal DT model, and the action model is updated to optimize the system. Numerical results also prove the viability of the algorithm.

With the help of edge computing techniques, MEC has become a novel approach for meeting the strict real-time requirements of DTs. Digital Twin Edge Networks (DITEN), which emerged as a novel DT architecture, have attracted significant attention. Sun et al. \cite{DT_07} first proposed a new vision of DITEN that considers user mobility and the variability of MEC environments, where Lyapunov optimization method and DRL algorithm are leveraged to handle the long-term migration cost. IoT is one of the most common application scenes of DT. For example, Lu et al. \cite{DT_08} integrated DTs with edge networks and proposed DITEN, while a blockchain-empowered federated learning scheme alleviates the data privacy protection and communication security problem in DITEN. A DT-assisted MEC model was proposed in \cite{DT_09} to minimize the end-to-end latency of industrial automation where multiple IoT devices and servers were considered. Additionally, UAVs can be applied as MEC nodes, providing low consumption and high flexibility for mobile-edge services. Li et al. \cite{DT_10} studied the intelligent task offloading problem in UAV-enabled MEC assisted by DT: Double Deep Q-network algorithm and closed-form expression method were deployed to optimize the system's energy consumption. A social-aware vehicular edge caching mechanism was designed in \cite{DT_11}, and a new concept of vehicular cache cloud was developed. The system utility is optimized by a deep deterministic policy gradient learning approach intelligently. In the Industrial IoT scenario, DITEN and federated learning were applied to construct DTs for IoT devices in the digital space. The energy and time cost of the communication process is minimized with the deep neural network model \cite{DT_12}.

From the above work, it is undeniable that MEC can improve the performance of DTs. Since the cloud servers are always far from the physical entities, the transmission in mobile networks will lead to higher latency for MCC. However, MCC still has its advantages. For example, the cloud server has higher computational power, and the highly integrated framework makes it more energy efficient. Therefore, the cloud server can provide a lower time and energy cost for entities with huge workloads during execution. Hence, more work should be done on integrating DTs with MEC and MCC heterogeneous environments. Moreover, current DT system models mainly focus on improving performance by reducing the total consumption of each DT in the system. However, the DT with multiple data resources has yet to be discussed. No consideration has been given to the synchronization problem between various devices when formula the system cost.

\subsection{Intelligent Offloading Schemes}

In addition to the DT system models, there is still plenty of work that mainly focuses on the offloading decision-making problem of MEC and MCC. Much of the current work is based on traditional optimization methods such as Lyapunov optimization \cite{Lya_01,Lya_02,Lya_03,Lya_04,Lya_05,Lya_06}, Graph Theory \cite{GRA_01,GRA_02}, Stalberg Game Theory \cite{F_SGA}, and Successive Convex Approximation \cite{NON_CONVEX}. These schemes can take optimal or near-optimal offloading decisions for various offloading environments. However, the problem complexity increased exponentially when the number of devices increased. Therefore, the time consumption during generating offloading decisions using traditional optimization methods will be unaffordable when dealing with complicated scenes. To better solve this NP-hard problem within a relatively short time, machine learning-based intelligent algorithms have been the main research direction in recent years.

Most machine learning algorithms require large amounts of data to train the networks. However, the optimal offloading decision data is challenging to obtain when facing dynamic heterogeneous MEC/MCC environments. So, algorithms that can be trained without data, such as RL and DRL based algorithms, have attracted much attention from many researchers. Huang et al. \cite{DRL_01} first proposed a Deep Q-Network based offloading algorithm to minimize the system cost in the MEC environment. In order to jointly optimize the task assignment and radio resource allocation in the MEC/MCC scenes, Dab et al. \cite{DRL_02} proposed the QL-JTAR algorithm based on Q-learning. Simulations conducted in NS3 have proved the performance of the approach. Double Deep Q-Network algorithm and a Q-function decomposition technique were combined in work \cite{DRL_03}. The problem of stochastic computation offloading is formulated as a Markov decision process, and two novel offloading algorithms for the MEC system, Deep-SARL and DARLING, are designed. When considering the application scenario of vehicular edge computing, which is a typical application of MEC, Zhan et al. \cite{DRL_04} presented a new DRL architecture based on proximal policy optimization method called DRLOSM. A convolutional neural network is also used to approximate policy and value functions and extract representative features. Qiu et al. \cite{DRL_05} studied a blockchain-based collective Q-learning approach in a networking integrated cloud–edge–end resource allocation in IoT.

RL and DRL-driven algorithms are promising solutions to problems that lack enough training data. However, another issue emerges when considering the dynamic environments in MEC/MCC: the model needs to be retrained when the offloading environment varies. Meta-learning is an excellent solution to improve the portability of different methods. Huang et al. \cite{META_01} combined Meta-learning with deep learning in the MEC network and proposed an algorithm called MELO. Through training the DNNs with historical MEC task scenarios, the model can achieve 99\% accuracy via one-step fine-tuning when facing a new MEC task scenario. Wang et al.\cite{META_02} first applied Meta-Reinforcement Learning in a multi-access edge computing network. They synergized the first-order MRL algorithm with a sequence-to-sequence neural network and proposed the MRLCO algorithm. The typical meta-learning algorithm MAML and Reptile are also applied in MRL and combined with DRL in \cite{META_03,META_04}. Simulations have proved that the training time can be remarkably reduced, and the model's portability can be improved significantly.

Apart from the meta-learning approach, distributed deep learning is another encouraging way to generate offloading decisions despite few training data. DDL algorithms use multiple parallel DNNs to generate offloading decisions. Each DNN receives offloading environment as input and output offloading decisions for each task through the forward propagation process. In contrast to traditional DNNs, which require large amounts of data for training, distributed deep learning can make the DNNs train each other without data. Huang et al. \cite{DDL_01} first proposed the DDLO algorithm based on DDL for the MEC networks. As a result, near-optimal offloading decisions can be generated in less than one second. More work about the online computation offloading issue in MEC networks was studied in \cite{DDL_02}. Thorough experiments and evaluation works proved the feasibility of the designed DROO algorithm. Moreover, Wu et al. \cite{DRL_03} extended the algorithm and proposed the distributed deep learning based offloading algorithm for the heterogeneous MEC/MCC system. Numerical results indicate that the average error can be controlled under 6\% compared to optimal decisions.

DDL has shown significant potential when facing MEC and/or MCC scenarios. Nevertheless, few works have been done for applying DDL in the DT system model. In addition, most approaches ignored the relationship between different devices during the deployment. The hidden information should be extracted and utilized to simplify the network structure. Here, we proposed a novel DT system model that considers the synchronization issues for the DT with various devices for data collection. The distributed deep learning algorithm is combined with a fully connected network layer to further extract the features from each group. The offloading decision-making model takes features as input instead of all the offloading environment information. 

\section{\scshape System Model and Problem Formulation}

This section elaborates on our DT system model framework of heterogeneous clouds, where each DT is maintained by one edge server or cloud server. Each DT is maintained via data collected from multiple devices, such as sensors and cameras. The data from different resources will be synchronized in the corresponding server. The optimization problem is then defined by formulating the system's total latency and energy consumption. For convenience, the major notations used in this paper are listed in Table~\ref{notation}.

\begin{table}[htbp]   
\renewcommand{\tablename}{TABLE}
\caption{Major Notations}   
\begin{center}
\resizebox{0.50\textwidth}{!}{%
\begin{tabular}{lll}    
\toprule    \bf{Notation} & \bf{Description}\\    
\midrule  $w_n$ & The size of data for the $n_{th}$ device\\ 
 $b_n$ & The bandwidth of the $n_{th}$ device \\  
 $f_c$ & The clock speed of CPU in the cloud server.\\
 $f_e^s$ & The clock speed of CPU in the $s_{th}$ edge server.\\
 $\gamma$ & The discount parameter of the network.\\
 ${\delta}$ & Instructions executed by CPU for one unit of data.\\
 $\theta_c$ & Average energy cost for each instruction in the cloud server. \\
 $\theta_e^s$ & Average energy cost for each instruction in the $s_{th}$ edge server. \\
$T^c_{t,n}$ & Transmission time cost when $w_n$ is executed in cloud server.\\
$T^c_{c,n}$ & Execution time cost when $w_n$ is executed in cloud server. \\
$T^e_{t,ns}$ & Transmission time cost when $w_n$ is executed in the $s_{th}$ edge server.\\
$T^e_{c,ns}$ & Execution time cost when $w_n$ is executed in the $s_{th}$ edge server.\\
$G_n$ & The ownership of the $n_{th}$ device.\\
$X_m$ & The placement of the $m_{th}$ DT.\\
$D_n$ & The offloading decision of $w_n$.\\
$\alpha$ & The weight coefficient between latency and energy consumption.\\

\bottomrule   
\end{tabular}  }
\end{center}
\label{notation}
\end{table}

\subsection{Framework}
\label{system_model}

\begin{figure*}[htbp]
	\centerline{
	\includegraphics[scale=0.165,trim=0in 0.1in 0in 0.4in]{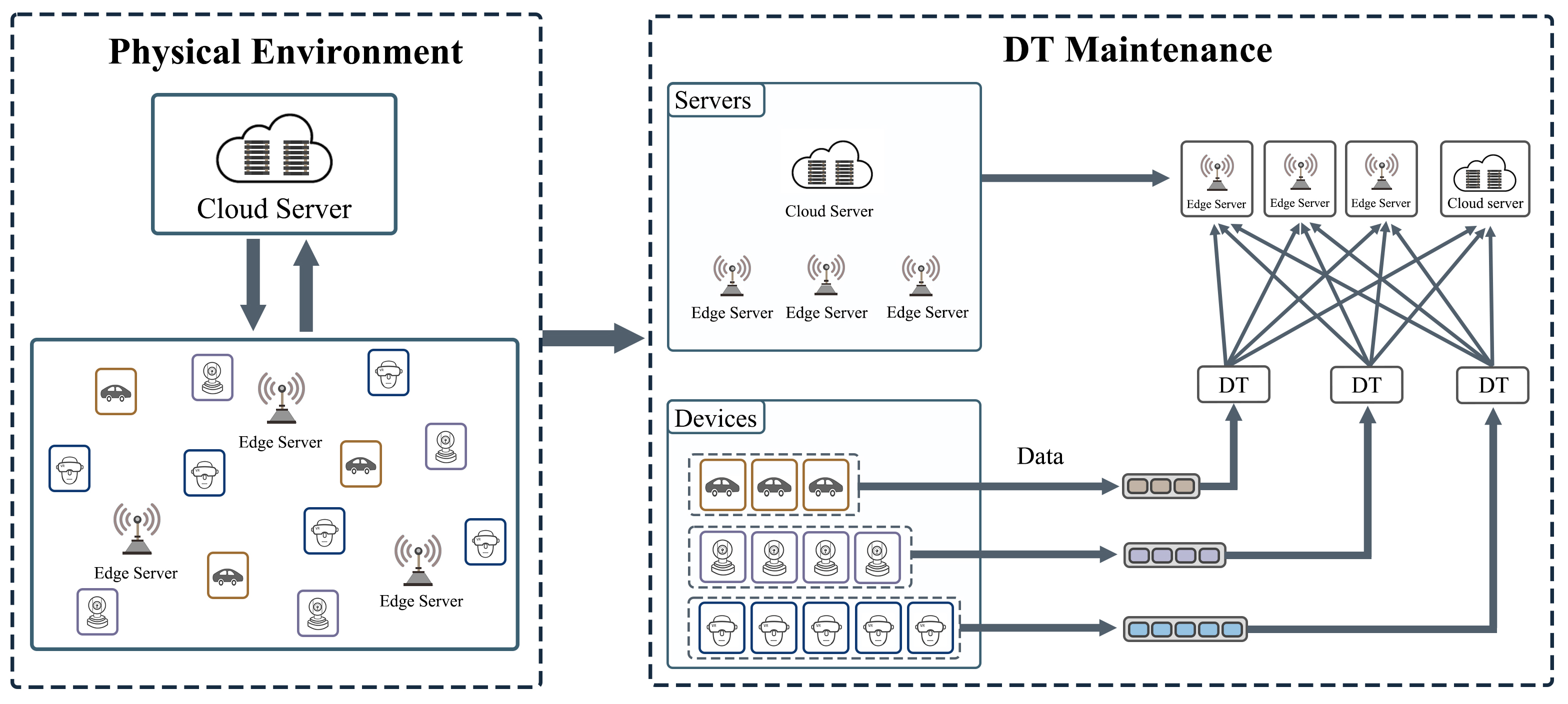}}
	\caption{Framework of the proposed DT system model}
	\label{system}
\end{figure*}

Figure \ref{system} elaborates the framework of our DT system model. The framework can be divided into two main parts: the physical environment and the procedure of DT maintenance. In the physical environment, we assume that there are multiple edge servers and various data collection devices in the field, where all the edge servers and devices are deployed. Each device can connect to the edge servers via wireless transmission, and one cloud server can also be utilized through the mobile network.

Besides, we assume that several DTs are maintained by the devices and servers in the physical environment. The maintenance of each DT requires data from various devices, i.e., the vehicles in the same street, Metaverse users in the same virtual space, and cameras for intelligent robots. Each DT is placed and maintained in one of the servers, where all the data is synchronized, and computational tasks are executed, such as simulation, prediction, and analysis.

DTs can achieve better performance with the advantages of MEC and MCC, but the placement of DTs in the dynamic scenario still matters. On the one hand, the utilization of computation power and bandwidth should be maximized in order to further reduce the delay; on the other hand, the average latency and energy consumption should be balanced accordingly. Therefore, the resource allocation problem during the DTs maintenance will be determined according to the environment.

\subsection{Offloading Model}

Our system model contains one cloud server, S edge servers, denoted as $\mathcal{S}=\{1,2,\cdots,S,S+1\}$, and various devices used for data collection, which can be denoted as $\mathcal{N}=\{1,2,\cdots,N\}$. Each device belongs to one of the DT and all the DTs can be represented as $\mathcal{M}=\{1,2,\cdots,M\}$. In addition, the maintenance of the DTs always requires great computational capacity, which is much higher than the capacity of the calculation unit in each device. The workload, including the newly collected information such as location and movement, is transmitted to the server via wired or wireless transmission techniques. The DTs are maintained in different servers instead of computing all the tasks locally. The whole process can be called offloading. We assume that the workload for the $n^{th}$ devices is $w_n$. The DTs can be maintained on either a cloud server or edge servers, and workloads are offloaded to the servers according to the offloading decision.

\subsubsection{MCC Model}

The cloud server is a powerful physical or virtual infrastructure that has powerful computational resources and great data storage capacity. By dividing a physical server into multiple virtual servers using virtualization approaches, users can buy cloud services according to the workload. Multiple users can be serviced simultaneously, and less energy is consumed due to the high integration and better utilization rate of the resources.

When the DT chooses to be maintained in the cloud server, the data from the $n^{th}$ devices is delivered via the mobile network. Since cloud servers are always far from the devices, we assume that the transmission time is only related to the data size and the allocated bandwidth. The transmission time can be formulated as follows:

\begin{equation}
    \label{cloud_trans_time}   
    T^c_{t,n}=\frac{w_n}{b_n\gamma}
 \end{equation}
where $b_n$ is the bandwidth between the $n^{th}$ device and the base station and $\gamma$ is the discount parameter considering the network fluctuations and congestion. Generally speaking, the time consumed during the download period can be ignored since only the necessary results are returned, which is much smaller than the size of offloaded data.

During the execution process, the clock speed of the CPU in the cloud server can be represented as $f_c$. And we assumed that ${\delta}$ instructions are executed when processing one unit of data. Thus the execution time can be denoted as follows:
\begin{equation}
    \label{cloud_cal_time}
    T^c_{c,n}=\frac{{\delta}{w_n}}{f_c}
 \end{equation}

To further evaluate the overall energy cost during cloud computing, the average energy cost for transmitting data of one unit size to the cloud server is set to $e^c_t$ and the average energy required to process each instruction is $\theta_c$. 
Then the total energy cost of $w_i$, which is consisted of the transmission energy and calculating energy, can be derived as :
\begin{equation}
    \label{cloud_energy}
    E^c_n=E^c_{t,n}+E^c_{c,n}={e^c_t}{w_n}+{\theta_c}{\delta}{w_n}
 \end{equation}

\subsubsection{MEC Model}
Compared to cloud servers, edge servers are placed near the users and devices in the same scenario. Benefiting from the short distance, the devices can directly communicate with the edge servers through wireless techniques such as WiFi and Bluetooth. The transmission speed is much faster in a more stable. However, limited by the radio distance, the transmission rate decreases as the distance grows, which makes distance one of the main factors affecting transmission performance. The latency during MCC transmission will be lower for devices that are far from edge servers.

To further define the latency, we use $L_n$ and $L_s$ to represent the location of the $n^{th}$ device and the $s^{th}$ edge server. The maximum transmission rate $p_{n,s}$ can be determined as:
\begin{equation}
    p_{ns}=\frac{\lambda}{||L_n-L_s||_2}
\end{equation}
where the distance between the device and the server is calculated by a $l_2$ norm. The $\lambda$ is the average discount parameter decided by the radio access distance of different devices and interference in the scenario. Then the transmission time cost when choosing the $s^{th}$ edge server can be derived as:
\begin{equation}
    \label{edge_trans_time}
    T^e_{t,ns}=\frac{w_n}{p_{ns}}=\frac{1}{\lambda}{w_n}{||L_n-L_s||_2}
\end{equation}

When executing the same task on cloud servers or edge servers, the number of instructions is the same, allowing the computation time to be expressed as:
\begin{equation}
    \label{edge_cal_time}
    T^e_{c,ns}=\frac{{\delta}{w_n}}{f_e^s}
\end{equation}
where $f_e^s$ represent the clock speed of CPU in the $s^{th}$ edge server.

Same as edge computing, the total energy consumption in cloud computing of $w_i$ can be determined as:

\begin{equation}
    \label{edge_energy}
    E^e_{n,s}=E^e_{t,ns}+E^e_{c,ns}={e^e_t}{w_n}+{\theta^s_e}{\delta}{w_n}
\end{equation}
where $e^e_t$ is the transmission energy cost parameter of MEC and $\theta^s_e$ is the execution energy parameter of the $s^{th}$ edge server.

\subsubsection{Problem Formulation}

Considering all the DTs and devices in the system, we use $\mathcal{G}=[G_1,G_2,\cdots,G_N]^\top$ to denote the ownership information of the devices. For the $n^{th}$ device, in details we have $G_n=[G_n^1,G_n^2,\cdots,G_n^M]$, which is denoted as: 

\begin{equation}
G_n^m= 
\begin{cases}
1, & \textrm{if the $n^{th}$ device belongs to the $m^{th}$ DT}, \\
0, & \textrm{otherwise}. 
\label{indicator1}
\end{cases}
\end{equation}

Meanwhile, we define that the offloading decision of the environment is $\mathcal{X}=[X_1,X_2,\cdots,X_M]^\top$ where ${X_m}=[X_{m,1},X_{m,2},\cdots,X_{m,S},X_{m,S+1}]$ can similarly be denoted as:

\begin{equation}
X_{m,s}= 
\begin{cases}
1, & \textrm{if the $m^{th}$ DT choose the $s^{th}$ server}, \\
0, & \textrm{otherwise}. 
\label{indicator2}
\end{cases}
\end{equation}
To be specific, it refers to the $s^{th}$ edge server when $1\leq s\leq S$ and the cloud server when $s=S+1$. 

Then the offloading decision of the devices can be derived as:
\begin{equation}
    \mathcal{D}=[D_1,D_2,\cdots,D_N]^\top={\mathcal{G}}{\times}{\mathcal{X}}
\end{equation}
 where $D_n$ is the offloading decision of the $n^{th}$ device.

The expected transmission time and calculation time of the $n^{th}$ entity under offloading decision $\mathcal{X}$ can be derived as:
\begin{equation}
    {T_n^t}={G_n}{\mathcal{X}}\psi_n^t
\end{equation}
\begin{equation}
    {T_n^c}={G_n}{\mathcal{X}}\psi_n^c
\end{equation}
 where $\psi_n^t=[T^e_{t,n1},T^e_{t,n2},\cdots,T^e_{t,nS},T^c_{t,n}]$ and $\psi_n^c=[T^e_{c,n1},T^e_{c,n2},\cdots,T^e_{c,nS},T^c_{c,n}]$ are the delay matrix according to the transmission time delay defined in (\ref{cloud_trans_time}) (\ref{edge_trans_time}) and calculation time delay defined in (\ref{cloud_cal_time}) (\ref{edge_cal_time}).

Similarly, we can define the energy consumption matrix as $\xi_n=[E^e_{n,1},E^e_{n,2},\cdots,E^e_{n,S},E^c_n]$ and the total energy cost of the $m^{th}$ DT according to (\ref{cloud_energy}) and (\ref{edge_energy}) can be written as:

\begin{equation}
    E_m=\sum_{n\in\mathcal{N},G_n=G_m} {G_n}{\mathcal{X}}\xi_n
\end{equation}

As a consequence, the total energy cost of the whole system can be written as:

\begin{equation}
    \label{system_energy}
    E=\sum_{m=1}^M (\sum_{n\in\mathcal{N},G_n=G_m} {G_n}{\mathcal{X}}\xi_n)
\end{equation}

Considering that the data synchronization process is done when all the devices of the same DT finish data transmitting procedure. Besides, the server can receive data from multiple devices simultaneously. So, the latency during synchronization is decided by the longest time cost of each DT, which can be expressed as:

\begin{equation}
    T_m^t =max\{T_n^t\big|n\in\mathcal{N};G_n=G_m\}
\end{equation}

Besides, we also need to take device number distinctions into account. Then the overall time cost of the group can be expressed as:
\begin{equation}
    T_m = \left|\left|I_m\right|\right|(T_m^t+\sum_{n\in\mathcal{N},G_n=G_m} T_n^c) \nonumber\\
\end{equation}
where $\left|\left|I_m\right|\right|$ denotes the devices number of the $m^{th}$ DT.

Therefore, the overall time and energy expenditure of the MEC/MCC hybrid offloading model can be derived as:
\begin{align}
    \label{system_time}
    T= &\sum_{m=1}^M (\left|\left|I_m\right|\right|(max\{{G_n}{\mathcal{X}}\psi_n^t\big|n\in\mathcal{N};G_n=G_m\}\nonumber\\
        &+\sum_{n\in\mathcal{N},G_n=G_m} {G_n}{\mathcal{X}}\psi_n^c))\nonumber\\
\end{align}

From the above equations, the system time cost defined in (\ref{system_time}) and system energy cost defined in (\ref{system_energy}) are related to the workload of each DT, the location of each device, the ownership of each device and the offloading decision of the model. In other words, for the giving offloading environment, we can calculate the system cost of different offloading decisions based on the above information. 

To find the optimal offloading decision considering the time and energy of the DT system, we further defined the weighted system cost function, which can be formulated as:
\begin{equation}
    Q(\mathcal{W},\mathcal{L},\mathcal{G},\mathcal{X})=\alpha T +(1-\alpha) E
    \label{cost_function}
\end{equation}
where $\mathcal{W} = \{w_n\big|n \in \mathcal{N} \}$ and $\alpha \in [0,1]$ is the weight coefficient to trade-off between energy consumption and total delay. For instance, system delay is the only concern when $\alpha=1$ and we only focus on the energy cost if $\alpha=0$.

Therefore, according to the DT network scenario, the optimal decision-making problem can be transformed into an optimization problem $\mathcal{P}$:

\begin{subequations}
\begin{equation}
    (\mathcal{P}) \quad min: Q(\mathcal{W},\mathcal{L},\mathcal{G},\mathcal{X})=\alpha T +(1-\alpha) E
\end{equation}
\begin{equation}
    s.t.: X_m^s \in \{ 0,1 \}, m \in \mathcal{M},s \in \mathcal{S}\qquad
\end{equation}
\begin{equation}
    \sum_{s=1}^{S+1} X_m^s=1, m \in \mathcal{M},s \in \mathcal{S}
\end{equation}
\end{subequations}

~\

The problem ($\mathcal{P}$) is a MIP and nonconvex problem with high-dimensional state space. In order to find the optimal solution for the nonlinear function, one needs to select the target decision from $(S+1)^M$ possible decisions. The computation complexity of the NP-hard problem increases exponentially when the DT and the server numbers become larger. In order to solve the problem efficiently, we developed a new approach for finding the near-optimal decision based on DDL.

\section{ALGORITHM}
\label{algorithm}

\begin{figure*}[htbp]
	\centerline{
	\includegraphics[scale=0.17,trim=0in 0.1in 0in 0.4in]{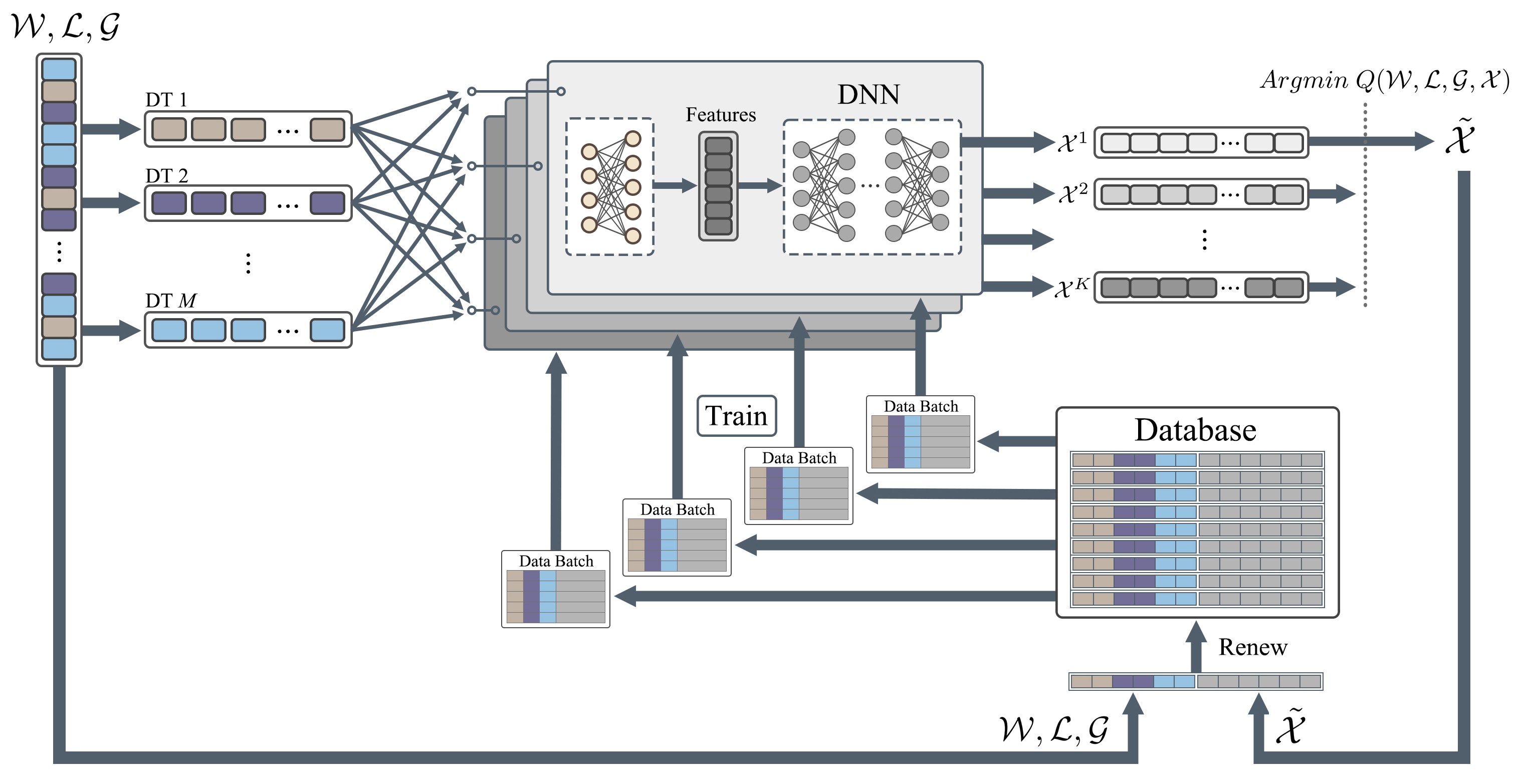}}
	\caption{The procedure of the proposed algorithm}
	\label{Frame_algorithm}
\end{figure*}

To solve the decision-making problem ($\mathcal{P}$) with efficiency and accuracy, we proposed a DDL based algorithm. A fully connected network, which can extract features of the workloads of each DT and simplify the network structure, is also added to solve the synchronization problem. The framework of our algorithm is depicted in Figure \ref{Frame_algorithm}.

\subsection{Framework}
Our framework can be divided into two parts, namely, the feature extract module based on a fully connected network and the decision module based on distributed deep learning approach. For any given workloads information $\mathcal{W}$, location $\mathcal{L}$ and devices information $\mathcal{G}$, we aim to output the respect near-optimal offloading decision $\mathcal{X}$.

Before the decision-making module, for each DT we first input $\mathcal{W}$ and $\mathcal{W}$ to the feature extractor module according to $\mathcal{G}$. Features for each DT are extracted, which then be regarded as inputs of the decision module to reduce the dimension of the problem. Although $\mathcal{W}$, $\mathcal{L}$ and $\mathcal{G}$ can be leveraged directly in the DDL process, the relationship between the devices of the same DT is ignored, and the number of trainable parameters is increased significantly, which will cause unnecessary computational cost. Besides, it is hard to use machine learning approaches to extract features. The training process always needs adequate training data, which is hard to get, and the features have to be pre-decided by experience. But in our scheme, the feature extraction module can be trained together with the decision module without data by utilizing the advantage of distributed deep learning, which will be elaborated in the next section.

The decision module consists of $\mathcal{K}$ parallel DNNs with the same structure but different initial weight parameters. Each DNN gets the data given by the feature extraction module as input and output offloading decisions of the application scenario. In the hidden layers, we use Relu as the activation function, and in the output layer, we use the Sigmoid activation function to map the output into (0,1). Since the offloading decision is represented by a sequence of binary integers, we have to map the values given by the network into the desired format. For the $k^{th}$ DNN, we use $\hat{\mathcal{V}}^k$ to represent the output values and $\mathcal{V}^k$ as the corresponding binary output. For any element $\hat{V}_i$ in $\hat{\mathcal{V}}^k$, the binary offloading decision $V_i$ can be determined by:

\begin{equation}
V_i= 
\begin{cases}
1, & \textrm{if $\hat{V}_i$ $\leq$ $\frac{1}{2}$}, \\
0, & \textrm{if $\hat{V}_i$ \textgreater $\frac{1}{2}$}. 
\label{decision_change}
\end{cases}
\end{equation}

As a result, The offloading decision $\mathcal{V}^k$ of each DT is represented by a sequence in a binary format. According to the definition of the offloading decision in problem ($\mathcal{P}$), each sequence is then converted from the binary format into one-hot format and composed into the required offloading decision $\mathcal{X}^k$. 

During the decision-making procedure, the input data is replicated and fed into each DNN. Since the multiple DNNs have different weights, we can obtain $k$ possible offloading decisions $\{\mathcal{X}^1, \mathcal{X}^2, \cdots, \mathcal{X}^K\}$. The performance in terms of time delay and energy consumption of each decision can be calculated by using the cost function $Q(\mathcal{W},\mathcal{L},\mathcal{G},\mathcal{X})$. Then the best decision $\Tilde{\mathcal{X}}$ with the lowest system cost can then be determined, which can be either used as the final decision or regarded as training data.

\subsection{Train}
Taking the complexity and variability of the MEC and MCC hybrid environments into account, we notice the difficulty of collecting abundant training data of the aiming scenario for the traditional deep learning based approaches. Besides, generating optimal decisions via conventional methods is also time-consuming when the server numbers and the DT numbers increase. In our approach, multiple DNNs are used to train the DDL model regardless of training data. The main procedure is depicted as follows.

We first randomly initialize the weight parameter of each DNN differently. To be specific, each layer obeys a normal distribution but the same layer of different DNNs have different weight parameter. Then, we randomly generate a set of input data consisting of $\mathcal{W}$, $\mathcal{L}$, and $\mathcal{G}$ to simulate one possible situation. After that, we input all the data to each DNN and these $\mathcal{K}$ parallel DNNs will output $\mathcal{K}$ different decisions. Based on the generated situation and decisions, we can calculate the current best decision $\Tilde{\mathcal{X}}$. Although the DNNs have not been trained well for the same input, the best decision still represents better performance, which can be regarded as labeled data. Improvements will be achieved theoretically if we use the selected decision to train other DNNs. Thus we regard $\{\mathcal{W},\mathcal{L},\mathcal{G},\Tilde{\mathcal{X}}\}$ as one training data and stored in the training data set. The process is iterated until the upper limit of data size is reached.

To train the model, we randomly choose a batch of training data from the data set for each DNN. The batches are different from each other since the weights of each neural network will soon become the same as the iteration progresses if all the DNNs are trained by the same data, which will be a hindrance to the exploration of possible decisions. Then cross-entropy loss function is applied for each DNN, which is minimized based on gradient descent. We denote the training data set of the $k^{th}$ DNN as $\mathcal{R}^k=\{(\mathcal{W}_i^k,\mathcal{L}_i^k,\mathcal{G}_i^k,\Tilde{\mathcal{X}}_i^k)\big |1\leq i \leq U\}$, where U is the size of each data batch. Then the loss of the $k^{th}$ DNN can be calculated as:

\begin{align}
    L(\pi_k)=&-\frac{1}{U} \sum_{i=1}^U ((\Tilde{\mathcal{X}}_i^k)^T {\log f}_{\pi_k} (\mathcal{W}_i^k,\mathcal{L}_i^k)\nonumber\\
    &+{(1-\Tilde{\mathcal{X}}_i^k)}^T \log(1-f_{\pi_k} (\mathcal{W}_i^k,\mathcal{L}_i^k)))
\end{align}
where $\pi_k$ to represent the parameter of the $k^{th}$ DNN and $f_{\pi_k}$ is the corresponding output.

After updating the parameter $ \{ \pi_1,\pi_2,\dots,\pi_K\}$ using Adam optimizer \cite{adam} according to the calculated loss value, the DNNs should have better performance and higher decision-making level. Therefore, we repeat the data generation process described in the previous part and replace part of the data set with these new samples that have higher accuracy. 

The parallel DNNs keep learning from the best decisions within each other, and the data set is constantly updated to ensure the training performance. Iterations are conducted, and the neural networks are continuously trained to approximate the global optimal offloading decisions. The pseudo-code of the proposed DRL algorithm is provided in \textbf{Algorithm~\ref{algorithm_code}}.

\begin{algorithm}[htbp]

\caption{DDL-based Offloading Algorithm} %
{\bf Input:} Workload $\mathcal{W}$, location $\mathcal{L}$, devices information $\mathcal{G}$\\
{\bf Output:} Offloading decision $\Tilde{\mathcal{X}}$
\begin{algorithmic}[1]
\State \textbf{Initialization:} Initialize the model with random parameter $\theta_1 $ and empty the database
\For {$j=1,2,3,\cdots,N$}  
    \State Randomly generate a group of input $\mathcal{W}_i$, $\mathcal{L}_i$ and $\mathcal{G}_i$
    \For {$i=1,2,3,\cdots,K$} 
        \State Input the generated data $\mathcal{W}_i$, $\mathcal{L}_i$ and $\mathcal{G}_i$ to the 
        \Statex \quad \quad \quad fully connected network;
        \State Input the extracted features to the $i^{th}$ DNN;
        \State Generate the $i^{th}$ offloading decision $\Tilde{\mathcal{X}}_i^k$;
    \EndFor
    \State Select decision $\Tilde{\mathcal{X}}_i=$ arg$\mathop{min}\limits_{\{\Tilde{\mathcal{X}}_i^k\}}$ $Q(\mathcal{W}_i,\mathcal{L}_i,\mathcal{G}_i,\Tilde{\mathcal{X}}_i^k)$;
    \State Calculate $Q(\mathcal{W}_i,\mathcal{L}_i,\mathcal{G}_i,\Tilde{\mathcal{X}}_i)$ as $Q_i$;
    \If{database is not full}
        \State Store $(\mathcal{W}_i,\mathcal{L}_i,\mathcal{G}_i,\Tilde{\mathcal{X}}_i,Q_i)$ into the database;
    \Else
        \State Replace the oldest data with $(\mathcal{W}_i,\mathcal{L}_i,\mathcal{G}_i,\Tilde{\mathcal{X}}_i,Q_i)$ 
        \State Randomly choose K batches of training data;
	\State Train each DNN using a selected batch of data
        \Statex \quad \quad \quad and update $\theta_j$ using the Adam optimizer;
    \EndIf
\EndFor 

\For {$k=1,2,3,\cdots,K$} 
    \State Input $\mathcal{W}$, $\mathcal{L}$ and $\mathcal{G}$ to the fully connected network;
    \State Input the extracted features to the $k^{th}$ DNN;
    \State Generate the $k^{th}$ offloading decision candidate $\Tilde{\mathcal{X}}^k$;
\EndFor
\State Select decision $\Tilde{\mathcal{X}}=$ arg$\mathop{min}\limits_{\{\Tilde{\mathcal{X}}^k\}}$ $Q(\mathcal{W},\mathcal{L},\mathcal{G},\Tilde{\mathcal{X}}^k)$;
\State \Return 
Offloading decisions $\Tilde{\mathcal{X}}$
\end{algorithmic}
\label{algorithm_code}
\end{algorithm}

\subsection{Test}

For traditional deep learning based approaches, the model's accuracy can be tested using the labeled data. However, the distributed deep learning based model is trained without labeled data. Besides, the optimal decision is hard to obtain using both traditional optimization methods and intelligent methods. To evaluate the performance and verify the feasibility of the proposed algorithm, we need to adopt a plausible verification method based on the characteristics of DDL. The testing process of the model can be separated into two parts, the convergence performance during training and the accuracy performance after the model is trained well.

Firstly, we need to ensure that the whole model converges after iterations. More specifically, after finite training iterations, the generated decisions of different inputs remain the same before and after extra iterations. Although the loss value can show the convergence performance of each DNN, the convergence of the model mainly focus on the best decision chosen from all the DNNs. So, we use $U_0$ randomly generated input data as the test data, which can be denoted as $\mathcal{R}^{\mathcal{T}}=\{(\mathcal{W}_i,\mathcal{L}_i,\mathcal{G}_i)\big |1\leq i \leq U_0\}$. We input $\mathcal{R}^{\mathcal{T}}$ into the model before and after each training epoch and generate the corresponding decisions. The system cost of each decision is then calculated, including the cost for the old decisions before training and the new decisions after training, which can be written as $C^{old}=\{C_i^{old}|1\leq i \leq U_0\}$ and $C^{new}=\{C_i^{new}|1\leq i \leq U_0\}$. The convergence rate can be derived as follows:
\begin{equation}
    \mathcal{C}=\frac{1}{U_0}\sum_{i=1}^{U_0}\frac{min(C_i^{old},C_i^{new})}{max(C_i^{old},C_i^{new})}
\end{equation}
where $\mathcal{C}\in(0,1]$ and the $\mathcal{C}$ will be close to one of the model convergent better.

After the model is proven to converge well, we still need to make sure that the model can make near-optimal decisions rather than converge in any local optimal solutions. So several baselines are applied to make comparison evaluations due to the lack of optimal decisions. we use the randomly generated dataset as the test data, too. For the same test data set, we will measure the average cost under given decision-making schemes and our approach, which can show the improvement and drawbacks of our approach. Besides, we also change the weight coefficient $\alpha$ into different value to ensure the proposed algorithm have reliable accuracy in various trade-off situations.


\section{PERFORMANCE EVALUATION}
\label{evaluation}
In this section, we set up numerical simulations under different application scenarios to demonstrate the effectiveness of the proposed approach. Experiments are carefully designed to evaluate the test standards presented above. 

\subsection{Simulation Setup}
In our simulation, we consider a heterogeneous edge/cloud environment consisting of one cloud server and three edge servers. Considering the system congestion and the number of threads, we set the equivalent clock frequency of the cloud server $f_c=3.5$ GHz. Similarly, the clock frequency of the edge servers is randomly distributed between $1.8\sim3.0$ GHz. In addition, since the cloud server has better energy efficiency than the edge servers, we set the calculation energy consumption parameter $\theta_c=0.1$ mJ and $\theta_e^s=0.125$ mJ, respectively.

Then we assume that the DT edge/cloud system contains up to 120 data collection devices, which belong to 15 different DTs at most. The locations and device ownership are randomly simulated in different situations, and the workload of each device is derived from the range $[10,40]$ MB. The transmission cost for one unit of data is set as $e_t^c=0.15$ mJ and $e_t^e=0.125$ mJ. Then we assume that the bandwidth between each device and base station is 1000 Mbps. The summary of our evaluation parameters is demonstrated in Table~\ref{settings}.

\begin{table}[htbp]   
\renewcommand{\tablename}{TABLE}
\caption{Evaluation Parameter}   
\begin{center}
\resizebox{0.49\textwidth}{!}{
\begin{tabular}{lll}    
\toprule    \bf{Parameters} & \bf{Values}\\    
\midrule  The number of cloud servers & 1\\ 
 The number of edge server & $\mathcal{S}$ = 3 \\  
 The number of devices for data collection & $\mathcal{N}$ = 120\\
 The number of DTs & $\mathcal{M}$ = 15\\
 The size of the field & 1000m $\times$ 800m\\
 The clock frequency of cloud server & $f_c=3.5$ GHz\\
 The clock frequency of edge server & $f_e^s=[1.8,3.0]$ GHz\\
 The data size of each device & $[10,40]$ MB. \\
The energy to transmit a unit of data in MEC & $e_t^e=0.125$ mJ\\
The energy to transmit a unit of data in MCC & $e_t^c=0.15$ mJ\\
The energy to execute each instruction in MEC & $\theta_c=0.1$ mJ\\
The energy to execute each instruction in MCC & $\theta_e^s=0.125$ mJ\\
The bandwidth of each device & $b_n$ = 1000 Mbps\\
\bottomrule   
\end{tabular}  }
\end{center}
\label{settings}
\end{table}

For each DNN in our model, we considered a network with two hidden layers as the decision-making module and a fully connected network with one hidden layer to extract the feature of each group. We implement the proposed algorithm in Python 3.8.0 with TensorFlow 2.9.1. All the simulations are performed based on an Intel Core i7-12700H CPU and 16 GB memory.

\subsection{Training Evaluation}

In this part, we measure the convergence performance under different learning rates, DNN numbers, and database sizes. We also generate a test set consisting of 1024 different scenarios as the test data set. After each iteration, we input the test set into the model and calculate the average cost after each iteration, which can represent the accuracy change of our approach.

\subsubsection{Impact of Learning Rate}
\begin{figure}[htbp]
	\centerline{
	\includegraphics[scale=0.25,trim=0in 0.1in 0in 0.4in]{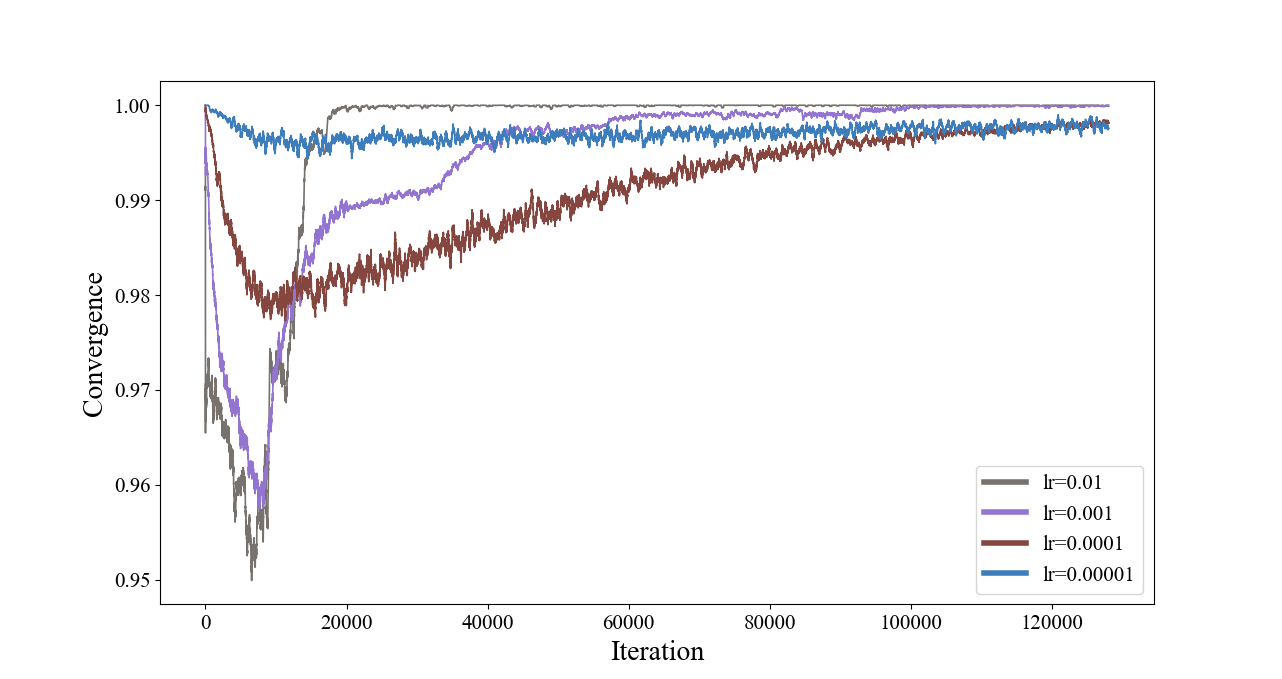}}
	\caption{Convergence performance of learning rates}
	\label{LR_c}
\end{figure}

\begin{figure}[htbp]
	\centerline{
	\includegraphics[scale=0.25,trim=0in 0.1in 0in 0.4in]{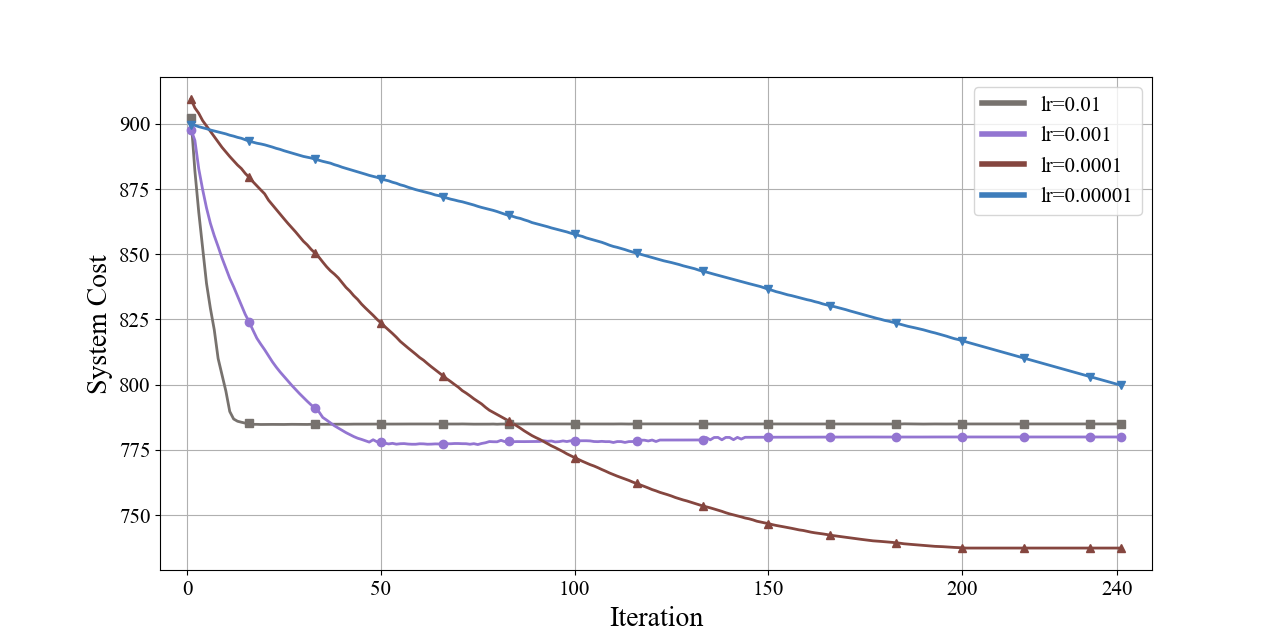}}
	\caption{Accuracy performance of learning rates}
	\label{LR_a}
\end{figure}

The learning rate is an essential hyperparameter that affects the performance of our approach. For the traditional DNN-based model, if the learning rate is too low, the model will converge very slowly, and more epochs are needed for training. On the contrary, the model will converge fast but cannot achieve good accuracy if the learning rate is too big.

In our experiments, we adjust the learning rate from 0.00001 to 0.01, and the convergence performance can be shown in Figure \ref{LR_c}. Since our DNNs are trained with the training data generated by the model itself, the impact of the learning rate is magnified by the iteration process. On the one hand, the parameters are changed too fast during training, and the database is replaced by the data with low accuracy, which will cause the model to skip a lot of hidden information and fall into a locally optimal solution or become under-fitting. From Figure \ref{LR_a}, we can see the accuracy under a high learning rate is unacceptable. On the other hand, the database is renewed at a low speed if the learning rate is too low. Thus the model is trained slower, which requires additional training costs.

\subsubsection{Impact of Number of DNNs}

\begin{figure}[htbp]
	\centerline{
	\includegraphics[scale=0.25,trim=0in 0.1in 0in 0.4in]{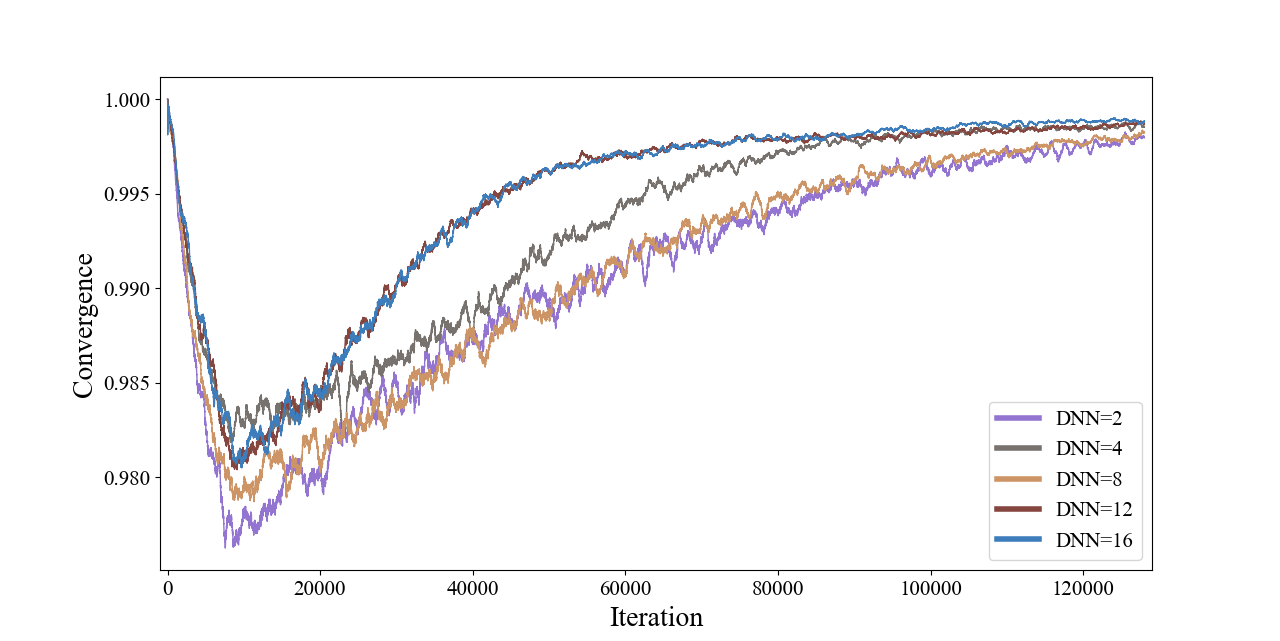}}
	\caption{Convergence performance of DNN number}
	\label{DNN_c}
\end{figure}

\begin{figure}[htbp]
	\centerline{
	\includegraphics[scale=0.25,trim=0in 0.1in 0in 0.4in]{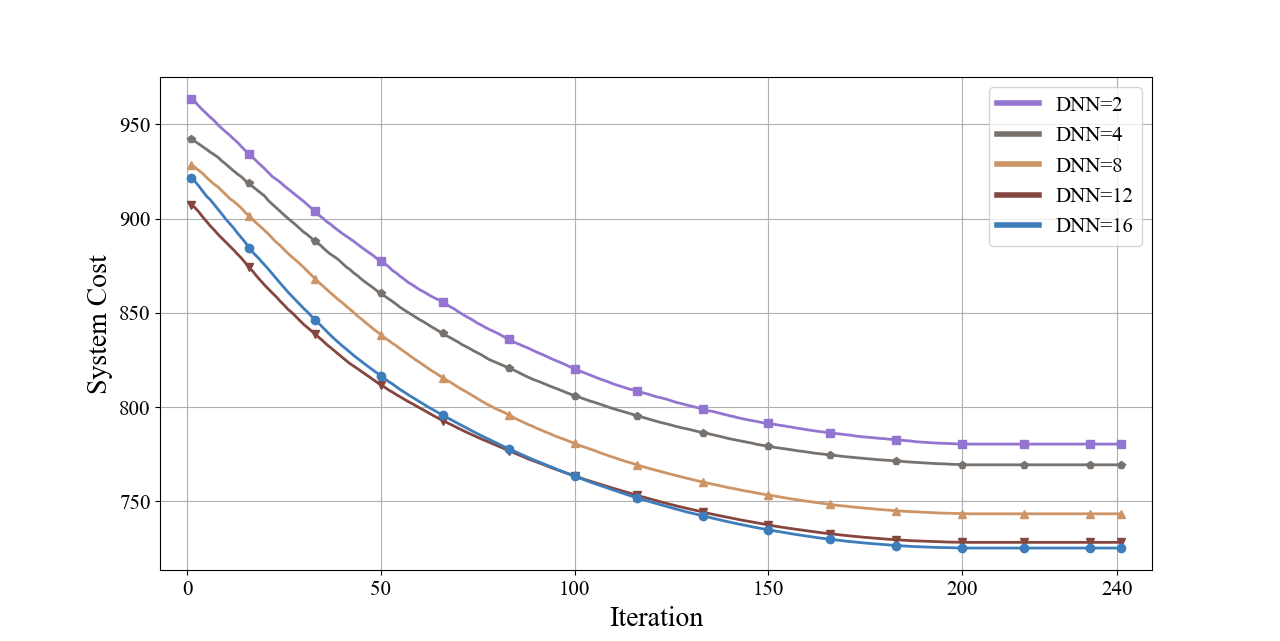}}
	\caption{Accuracy performance of DNN number}
	\label{DNN_a}
\end{figure}

During our experiments, the number of DNNs is adjusted from 2 to 16, and each model is trained with 1000 iterations. The convergence rate and accuracy performance are measured after every iteration using the approach defined in the previous section. From Figure \ref{DNN_c}, we can see that the model converges with the training iterations goes. The changing speed became faster at the beginning and then became stable since the decision is not changed until the output number crosses the dividing line according to equation \ref{decision_change}.

When the number of DNNs increases, fewer iterations are needed for the model to converge and the stability during training increases. From Figure \ref{DNN_a}, we can also derive that more DNNs can have better initial performance and higher accuracy can be achieved. However, additional DNNs can not ensure better accuracy and convergence speed when the model has used enough DNNs. As a consequence, we use 12 DNNs to avoid redundant training costs.

\subsubsection{Impact of Database Size}

Sufficient training data is needed when training a traditional DNN model. When the data is inadequate, the model will have low accuracy and may become over-fitting or under-fitting. To ensure the ability of the model to explore more possible decisions, in our model, we train different DNNs with different batches of datasets, and part of the database is renewed after each iteration.

\begin{figure}[htbp]
	\centerline{
	\includegraphics[scale=0.25,trim=0in 0.1in 0in 0.4in]{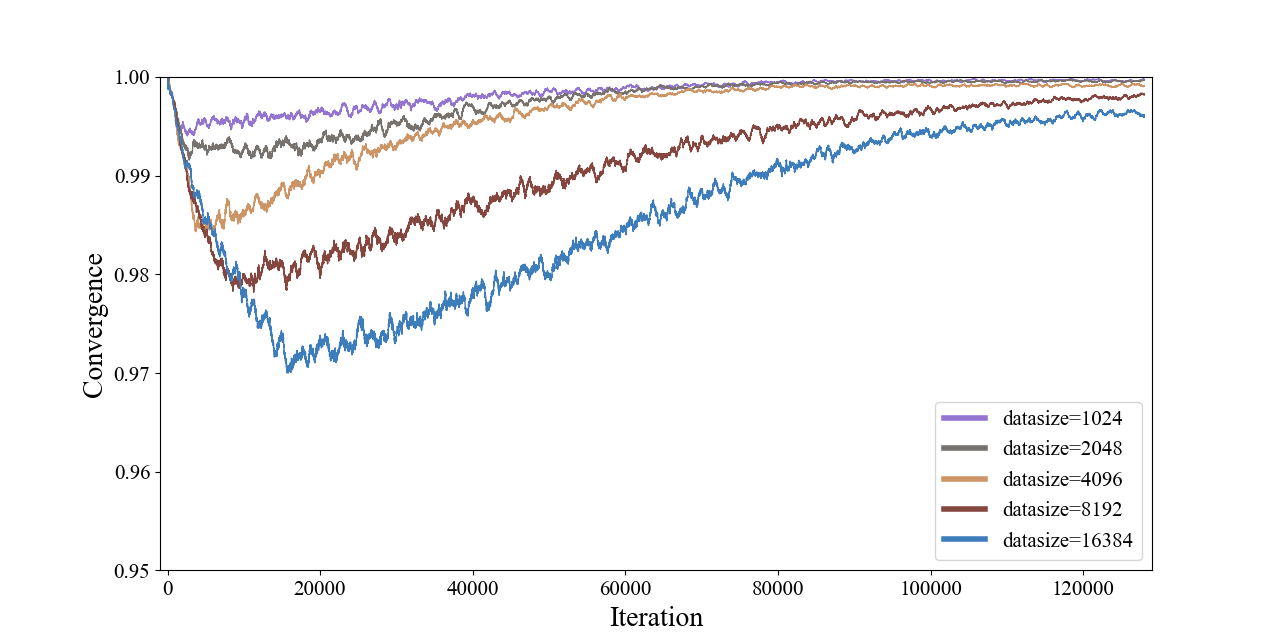}}
	\caption{Convergence performance of database size}
	\label{datasize_c}
\end{figure}

\begin{figure}[htbp]
	\centerline{
	\includegraphics[scale=0.25,trim=0in 0.1in 0in 0.4in]{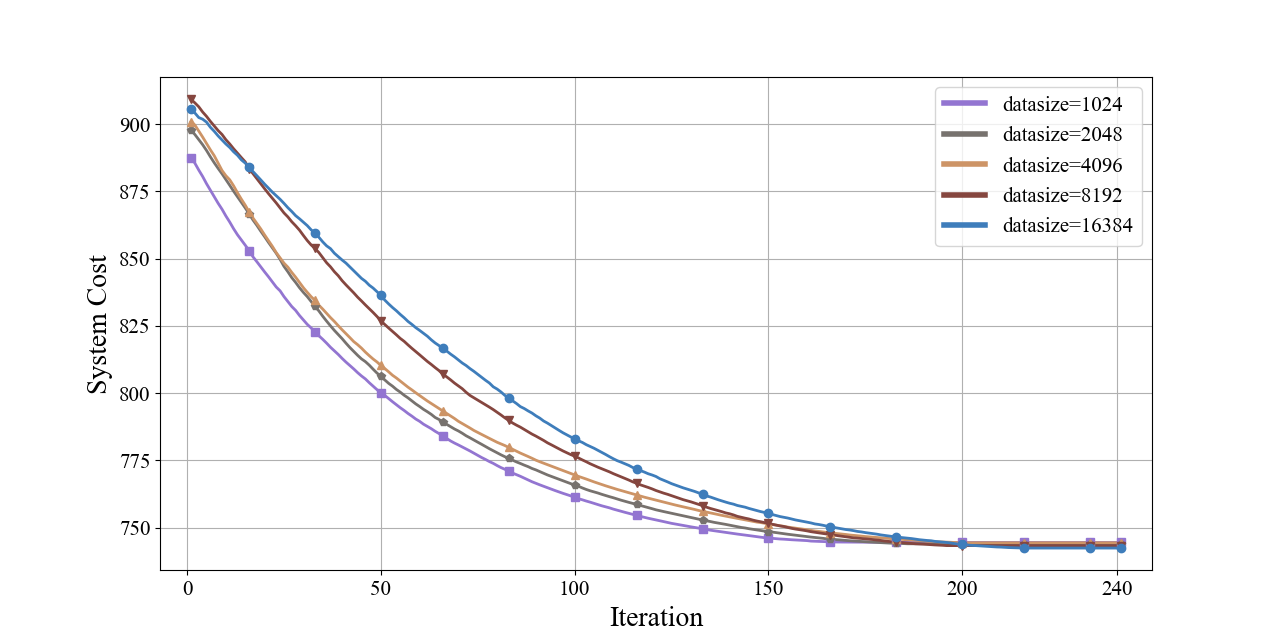}}
	\caption{Accuracy performance of database size}
	\label{datasize_a}
\end{figure}

As shown in Figure \ref{datasize_c} and \ref{datasize_a}, the database size has little impact on the decision-making model compared with the learning rate and number of DNNs. When the database size is sufficient for training iteration, extra data will not improve the accuracy of the model. The database is renewed slower, and a relatively longer training time is needed. So when facing a new application environment, we can use a database with a higher volume first to test the feasibility of the model and then decrease the size to minimize the training cost.

\subsection{Performance Comparison}

\begin{figure}[htbp]
	\centerline{
	\includegraphics[scale=0.3,trim=0in 0.1in 0in 0.4in]{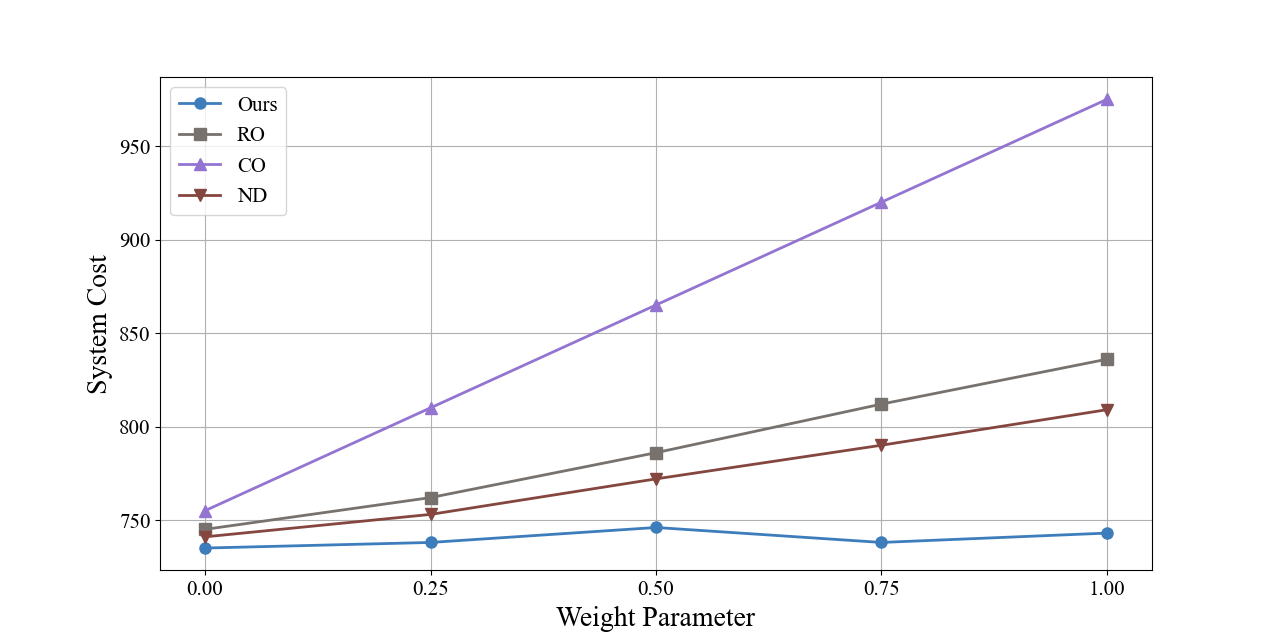}}
	\caption{Comparison with various schemes under different weighting parameters}
	\label{compare}
\end{figure}

By evaluating the convergence and accuracy performance under different hyperparameter settings, we have proven that the proposed algorithm can be trained without labeled data and significantly reduce system consumption. To have a better insight of the effectiveness of our approach, we choose several traditional schemes as baselines. We then change the weight coefficient from 0 to 1 and test the average system cost of different schemes. The following schemes are tested for comparison analysis:

\begin{itemize}
\item \emph{Random Offloading(RO)}: In this method, all the decisions in the decision space are chosen randomly with the same probability.
\item \emph{Cloud Only(CO)}: In this method, all the DTs are maintained by the cloud server.
\item \emph{Average Distribution(AD)}: In this method, all the DT is divided according to the size of the workload. Each server shares an average amount of workload.
\item \emph{Our algorithm}: In this method, we applied the proposed algorithm to generate offloading decisions.
 \end{itemize}

The comparison results of different offloading schemes are depicted in Figure \ref{compare}. The weight coefficient $\alpha$ of the cost function (\ref{cost_function}) represents the trade-off between system delay and total energy consumption. It can be seen that our model can converge and decrease the system cost significantly under different weight coefficients. Compared with several baselines, our approach can achieve better performance, especially when tackling the system delay. Besides, our algorithm also has better capabilities to handle wider application scenarios. We do not need to train the model again when the scenario changes, e.g., when each device's workload and location information are different. The decision can be obtained through a forward propagation process, which can be achieved within 0.01s.

\section{CONCLUSION AND FUTURE WORK}
\label{conclusion}

In this paper, we designed a new DT system model based on cloud computing and edge computing technology. In the model, we assume each DT have multiple data resource, and one server is chosen to synchronize the data and maintain the DT. To optimize the system cost and allocate the servers properly, we convert the resource allocation problem into a MIP problem. A new algorithm is then proposed based on DDL to generate near-optimal decisions intelligently. Simulation shows that our model can be trained according to the cost function and converge with good performance. No labeled data is needed for the network training, which allows our model to be used in a broader range of cases. In the given environment, the trained model can make decisions for different workloads and device locations with portability and efficiency. According to the comparison analysis, the energy consumption and total delay are significantly decreased compared with various baselines.

Although our work only considers a basic DT edge/cloud system model and various variables are ignored, we mainly focus on proving the feasibility of the proposed algorithm. This model is highly prospective and can be easily expanded into complicated real-world scenarios by changing the cost function. In the future, we will do more study about applying our algorithm to other traditional transmission models. To address the problem more effectively in applications, it is essential to evaluate the performance of both meta-learning and deep reinforcement learning within the same model to determine the most efficient approach. Besides, privacy protection and user data security are other issues of the model because of the potential risks associated with its collection, storage, and usage. Since the distributed deep learning based algorithm is very compatible with federal learning and blockchain technologies, more work should be done to ensure the system's privacy.

\ifCLASSOPTIONcaptionsoff
  \newpage
\fi



%
\bibliographystyle{IEEEtran}
\bibliography{reference.bib}

\end{document}